\documentclass[runningheads]{llncs}

\usepackage[utf8]{inputenc}
\usepackage{booktabs} 
\usepackage{color}
\usepackage{graphicx}
\usepackage{float}
\usepackage[ruled]{algorithm2e} 
\usepackage{amsmath}
\usepackage{amsfonts}
\usepackage{array}
\usepackage{caption} 

\newcommand{\aex} {adversarial example}
\newcommand{\aexs} {adversarial examples}

\newcommand{\Drebinfive} {$\text{Drebin}_{500}$}
\newcommand{\apds} {APDS}

\begin{document}

\title{Effectiveness of Adversarial Examples and Defenses for Malware Classification}
\titlerunning{Adversarial Examples and Defenses for Malware Classification}
\author{Robert Podschwadt \inst{1} \and Hassan Takabi\inst{1}}

\authorrunning{R. Podschwadt \and H. Takabi}
\institute{University of North Texas, Denton TX 76201, USA 
\email{robertpodschwadt@my.unt.edu} ,\email{takabi@unt.edu}}

\maketitle

\begin{abstract}
Artificial neural networks have been successfully used for many different classification tasks including malware detection and distinguishing between malicious and non-malicious programs. Although artificial neural networks perform very well on these tasks, they are also vulnerable to adversarial examples. An adversarial example is a sample that has minor modifications made to it so that the neural network misclassifies it. 
Many techniques have been proposed, both for crafting adversarial examples and for hardening neural networks against them. Most previous work has been done in the image domain. Some of the attacks have been adopted to work in the malware domain which typically deals with binary feature vectors. 
In order to better understand the space of adversarial examples in malware classification, we study different approaches of crafting adversarial examples and defense techniques in the malware domain and compare their effectiveness on multiple datasets. 


\end{abstract}

\section{Introduction}
Machine learning and deep neural networks in particular have been highly successful in different applications. They can be used for image recognition with human accuracy or above \cite{DBLP:journals/corr/abs-1708-04552}\cite{DBLP:journals/corr/ZagoruykoK16}, and they have also been employed to perform malware detection \cite{1710.09435}\cite{7090993}\cite{7413680}. 
A problem with neural networks is that they have been found to be vulnerable to adversarial examples created by perturbing input data.
Most adversarial machine learning research \cite{1412.6572}\cite{DBLP:journals/corr/CarliniW17}\cite{DBLP:journals/corr/Moosavi-Dezfooli15}\cite{DBLP:journals/corr/KurakinGB16} focuses on images since it is easy to visualize the perturbations, and all changes made to an image still produce a valid image. When perturbing malware instead of images \cite{DBLP:journals/corr/GrossePM0M16}\cite{DBLP:journals/corr/ChenXFHXZ17}, there are more constraints that need to be taken into consideration. This paper investigates neural networks applied to malware classification. 
We focus on a range of malware attacks and defenses proposed in recent years, providing an overview of the algorithms and comparing their efficacy in a consistent manner.
With an increasing number of attacks and defenses available, it is non-trivial to compare their effectiveness. Most publications use different datasets and different model architectures which makes it even harder to compare different approaches. Currently there is no straightforward way of telling which defense is efficient against which attack and vice versa. In this paper, we compare twelve attacks, their effectiveness against undefended models, and the impact of four proposed defenses. To achieve this, we use two different datasets from the malware domain and train deep neural networks on both. We implement or use existing implementations of the attacks and defenses to gain a better understanding of which defenses are strong in the malware domain. Our results show that all defenses, except for adversarial training, perform worse on malware than they do on images. 

\section{Background}

\subsection{Threat Model}

In a malware detection setting, the system's goal is to accurately distinguish between malware and non-malware. The attacker in this case is the author of a malicious program, and their goal is to evade detection by the malware dection system. In order to evade the detection system the attacker has two options. The first option is to ``sabotage" the classifier to the point that it becomes untrustworthy; the second is to modify their software is such a way that the classifier views it as benign. It is possible that the target of an attacker is to have a benign sample classified as malicious but it is a less straightforward attack like flooding the classifier with false-positives therefore eroding the users trust in the system.

While neural networks bring advantages for malware detection, their use leaves the system vulnerable to attacks that can be leveraged against neural networks in general.
If an attacker can mislabel instances that are in the training data it is called a data poisoning attack \cite{1206.6389}\cite{Huang:2011:AML:2046684.2046692}. The goal of data poisoning is to evade the classifier by having it trained on mislabeled data.
In this paper, we consider a less powerful attacker who can only interact with the system after it has been trained and has no influence over the training data. All that they can do is present an instance to the detection system and receive the classification result. For most methods of creating adversarial examples, the attacker needs access to the internal structure and information of the model. But even if the attacker only has query access to the system, attacks are still possible. Papernot et al. \cite{DBLP:journals/corr/PapernotMGJCS16} have shown that any black box attack can be transformed into a semi white box attack. This is due to transferability. It has been shown that adversarial examples created on one model can also be used to attack another model as long as the models have learned a similar enough decision boundary. 
For this reason, attacks in this paper are all white box attacks.

\subsection{Adversarial Examples}
First described by Szedgy et al. \cite{DBLP:journals/corr/SzegedyZSBEGF13} \aexs{} are instances that are very close to a legitimate instance but are classified differently. Give an instance $x$ with the class label $C(x)=t$ one can find an instance  $x'$ such that $ C(x') \ne t$. An additional constraint is that some distance metric $||x-x'||$ should be kept small. This means the changes made to the legitimate sample should be as small as possible. In images that usually means that it is impossible for a human to perceive the changes between $x$ and $x'$ with the naked eye. 
Different norms including $L_0$ norm \cite{su2017one} \cite{DBLP:journals/corr/CarliniW16a}, $L_1$ norm \cite{carlini2017ground}, $L_2$ norm \cite{DBLP:journals/corr/CarliniW16a}, and $L_\infty$ \cite{warde201611} have been used.
Since we are working with binary feature indication vectors, feature values can either be 0 or 1. Therefore the $L_\infty$ norm  would always be 1 if at least one change was made. The $||x-x'||_0$ norm yields all the changes that have been made. This makes $L_0$ a good choice.

\section{Experiments}

\subsection{Data}
We use the Drebin dataset \cite{drebin} of Android apps. It contains the extracted features for both the malware and the benign class. Additionally it also contains the 5560 malware apps themselves. Drebin relies on static features, i.e., those that can be extracted from an app without executing it. The specific features are: requested hardware, requested permissions, app components, filtered intents, restricted API calls, used permissions, suspicious API calls and network addresses. Those features come either from the manifest or decompiling the code.

We also run experiments on another dataset \apds{}, which consists of permissions extracted from android apps. The data is available on \textit{kaggle} \cite{kaggle}. It contains 398 instances with 330 features. 50\% of the samples are malware and 50\% are benign. 

\subsection{The victim model}
To test the performance of different attacks and defenses we define a victim model. The model we use consist of four fully connected hidden layers with Rectified Linear Unit (ReLU) as activation function. The architecture of the victim model can be found in Table \ref{victim_model_table}. The input layer is omitted since it depends on the dataset. We trained the model for 100 epochs using the Adam optimizer with a learning rate of .001, $\beta_1=.9$, $\beta_2=.999$. As a loss function binary cross entropy is used. The performance of the victim model can be found in Table \ref{victim_metrics_table}. For all of our training, we use 56.25\% of the data as training data, 18.75\% as validation data and 25\% as test data.   

\begin{table}
	\centering
	\caption{Accuracy (acc) of the \textit{victim model} on the testset. Also the false negatives rate (FNR), false positives rate (FPR), true positives rate (FPR) and true negatives rate (TNR) }
	\begin{tabular}{|c|c|c|c|c|c|} 
		\hline
		Dataset 		& acc  & FNR  & FPR  & TPR  & TNR  \\ 
		\hline

		$\text{Drebin}_{500}$	& 0.99 & 0.10 & 0.01 & 0.89 & 0.99 \\ 
		\apds 			& 0.98 & 0.04 & 0.0  & 0.96 & 1.0  \\ 
		\hline
	\end{tabular}
	\label{victim_metrics_table}
\end{table}

\begin{table}
	\centering
	\caption{Architecture of the victim model}
	\begin{tabular}{|c| c c|} 
		\hline
		Layer & Neurons & activation function \\
		\hline
		Fully connected & 300 & ReLU \\ 
		Fully connected & 250 & ReLU \\
		Fully connected & 200 & ReLU \\
		Fully connected & 128 & ReLU \\
		Fully connected & 2 & softmax \\
		\hline
	\end{tabular}

	\label{victim_model_table}
\end{table}

\subsection{Feature Selection} 

The Drebin dataset is composed of features for over 120,000 instances of which roughly 5,500 are malware. Each instance has approximately 550,000 features. The size of the instances makes the dataset unwieldy to use, and building a neural network graph with features that size leads to memory problems very quickly. Grosse et al. \cite{DBLP:journals/corr/GrossePM0M16} find that only a small number of features are used for creating adversarial examples. In order to test more complex models without running into memory problems, we employ feature selection on the victim model (Table \ref{victim_model_table}). We use the scikit-learn \cite{scikit-learn} implementation of SelectKBest with chi-squared as the scoring function and run feature selection for different values of K. The accuracy on the benign class does not change much with the number of features. This is due to the class imbalance. With 500 and 1000 features we achieve 99.6\% accuracy on the benign class and 89.2\% on the malware class.
Throughout the paper, when we want to indicate that we have used feature selection, we will indicate the number of features K that we have used as follows: $\text{Drebin}_K$.

\section{Attacks}

\subsection{JSMA}
Grosse et al. \cite{DBLP:journals/corr/GrossePM0M16} adapted the approach developed for images in  \cite{DBLP:journals/corr/PapernotMJFCS15} based on Jacobian Saliency Maps (JSMA). The Jacobian matrix is the matrix of the forwards derivatives for every feature of a sample wrt the output classes. The creation of \aexs{} consists of two steps. First the Jacobian is calculated. In the second step the feature with maximum positive gradient towards the desired output class is selected and perpetuated. In order to guarantee a correctly functioning \aex{} the constraint to only add features and not remove any is added. The steps of calculating the Jacobian and changing a feature are repeated multiple times, until a maximum number of allowed perturbations has been made or misclassification is achieved.  For a more in depth description see Algorithm 1 in \cite{DBLP:journals/corr/GrossePM0M16}.

The JMSA attack is effective against our trained victim models (Table \ref{attacks_grosse}), with evasion rates varying depending on the data set. While on \apds{} an evasion rate of 100\% can be achieved on the $\text{Drebin}_{500}$ data we were only able to evade the classifier 33\% of the time.
The change column gives the average changes made to all adversarial examples. 

\begin{table}
	\centering
	\caption{Grosse et. al \cite{DBLP:journals/corr/GrossePM0M16} approach based on JSMA with a maximum of 25 allowed perturbations.  }
	\begin{tabular}{|c|c|c|c|c|c|c|} 
		\hline
		Dataset 		& acc & acc adv & FNR  & FNR adv  & \# Features changed & evasion\\ [0.5ex] 
		\hline
		$\text{Drebin}_{500}$	& 0.99 & 0.98   & 0.01 & 0.19     & 1.89  & 0.33 \\ 
		\apds 			& 0.93 & 0.63   & 0.12 & 0.56     & 13.57 & 1.00  \\ 
		\hline
	\end{tabular}

	\label{attacks_grosse}
\end{table}

\subsection{Feature enabling and disabling} 
Stokes et al. \cite{DBLP:journals/corr/abs-1712-05919} propose different variants for creating \aexs{}, relaxing the constraints on allowed modifications. Instead of only adding features, which preserves the functionality of the malware, the authors also allow for the removal of features. The reasoning is that attackers can find different ways to implement the same malicious behavior. While this might be true, it requires a great deal of effort and is nowhere near as straightforward as only allowing the addition of features. We include the methods for completeness.

Stokes et al. propose three iterative methods which are based on the Jacobian using positive features and negative features. A positive feature is a feature that is an indicator of malware, meaning that the Jacobian for the malware class with respect to the input is positive for this feature. Similarly, a negative feature is an indicator for the benign class. From these attributes. From these attributes, the authors develop three different iterative methods for creating \aexs{}. For every iteration the Jacobian of the sample is calculated wrt. the output classes and one feature is perpetuated. This is done until the maximum number of allowed perturbations is reached or the sample is misclassified. Iteratively disabling positive features is called the dec\_pos attack, and disabling negative features, inc\_neg. These approaches can be applied in alternating fashion to get dec\_pos + inc\_neg. Instead of choosing the feature with the maximum value, a random feature can be chosen. This leads to three more different techniques; randomized dec\_pos, randomized inc\_neg and randomized dec\_pos + inc\_neg. Note that inc\_neg is the same strategy as that described in \cite{DBLP:journals/corr/GrossePM0M16}. We apply the attacks to our victim model trained on \Drebinfive{} the results can be seen in Table \ref{attacks-dec-pos-inc-neg}.

\begin{table}
	\centering
	\caption{ \cite{DBLP:journals/corr/abs-1712-05919} approaches, enabling and disabling features based on their Jacobian. Numbers reported on $\text{Drebin}_{500}$ }
	\begin{tabular}{|l|c|c|c|c|c|c|} 
		\hline
		Attack						& acc & acc adv & FNR  & FNR adv  & change & evasion\\
		\hline
		\hline
		dec\_pos					& 0.99 & 0.98   & 0.05 & 0.16     & 1.65  & 0.25 \\ 
		\hline
		inc\_neg 					& 0.99 & 0.98   & 0.05 & 0.19     & 1.89  & 0.32 \\ 
		\hline
		dec\_pos inc\_neg			& 0.99 & 0.98   & 0.05 & 0.16     & 0.08  & 0.27 \\
		\hline
		rand. dec\_pos				& 0.99 & 0.98   & 0.05 & 0.15     & 2.09  & 0.24 \\
		\hline
		rand. inc\_neg      		& 0.99 & 0.99   & 0.05 & 0.06     & 0.09  & 0.06 \\
		\hline
		rand dec\_pos inc\_neg	& 0.99 & 0.99   & 0.05 & 0.12     & 0.23  & 0.18 \\		
		\hline
	\end{tabular}

	\label{attacks-dec-pos-inc-neg}
\end{table}

randomized inc\_neg achieves no additional evasions at all, suggesting that randomly enabling features is not a good approach. The relative effectiveness of the attacks is similar over all compared models. The notable difference is that on \Drebinfive{} inc\_neg achieves a higher evasion with only slightly increased average change when compared to dec\_pos. On the other victim model, dec\_pos is able to achieve the same evasion rate as inc\_neg although at a smaller change cost.

\subsection{FGSM with rounding and Bit ascending}
Huang et al.  \cite{DBLP:journals/corr/abs-1801-02950} propose four different approaches. The first two methods are based on the \textit{Fast Gradient Sign Method}. FGSM is a one step method of moving the sample in the direction of the greatest loss. A more powerful variation of the attack is multi step $\text{FGSM}^k$ where $k$ is the number of iterations. To apply this method to the discrete malware domain rounding is required. The authors propose two different rounding schemes: deterministic rounding $\text{dFGSM}^k$ and random rounding $\text{rFGSM}^k$.
A third method is introduced which visits multiple promising vertices and explores the possible feature space more thoroughly. This method is called multi-step Bit Gradient Ascent($\text{BGA}^k$). $\text{BGA}^k$ works by setting a bit if the partial derivative of the loss is greater or equal to the $l_2$ norm of the loss gradient divided by $\sqrt{m}$. The reasoning of the authors is that those features contribute more or equally to the $l_2$ norm of the loss. 
Another method introduced is multi-step Bit Coordinate Ascent $\text{BCA}^k$. At every step the bit that corresponds to the feature with the maximum partial derivative of the loss is changed. 
The results of attacking the model trained on \Drebinfive{} can be found in Table \ref{attacks_fgsm_and_bit_ascent}. \begin{table}
	\centering
	\caption{ \cite{DBLP:journals/corr/abs-1801-02950} approaches $FGSM^k$, $rFGSM^k$, $dFGSM^k$, $BGA^k$ and  $BCA^k$. Numbers reported on $\text{Drebin}_{500} $ }
	\begin{tabular}{|c|c|c|c|c|c|c|} 
		\hline
		Attack		& acc & acc adv & FNR  & FNR adv  & change & evasion\\ 
		\hline
		\hline
		$dFGSM^k$	& 0.99 & 0.99   & 0.05 & 0.10     & 7.16  & 0.15 \\ 
		\hline
		$rFGSM^k$ 	& 0.99 & 0.99   & 0.05 & 0.10     & 7.16  & 0.15 \\ 
		\hline
		$BGA^k$		& 0.99 & 0.99   & 0.05 & 0.05     & 1.14  & 0.05 \\
		\hline
		$BCA^k$		& 0.99 & 0.99   & 0.05 & 0.09     & 1.26  & 0.12 \\
		\hline
	\end{tabular}

	\label{attacks_fgsm_and_bit_ascent}
\end{table}

The iterative variants of FGSM produce the most effective \aexs{} when it comes to evasion. The rounding scheme employed matters little. The same pattern holds true when we attack the other victim model. On the \apds{} model both attacks achieve 100\% evasion rate. $\text{BGA}^k$ fails to achieve any significant evasion beyond the natural loss across all models, while $\text{BCA}^k$ performs very different depending on the model. On \Drebinfive{} $ \text{BCA}^k$ achieves 12\% evasion rate which is almost 2.5 times the natural evasion rate of 5\%. On \apds{} the evasion rate achieved by the attack is 14\% which is barely more than the natural rate of 12\%.

\subsection{MalGAN}  

Malgan is an attack introduced by Hu et al. \cite{DBLP:journals/corr/HuT17}. In order to create adversarial examples it uses Generative Adversarial Network (GAN \cite{DBLP:conf/nips/GoodfellowPMXWOCB14}). The idea behind GANs is to create data by training an artificial neural network. For our experiment we use the architecture described by Hu et al. The Generator takes a legitimate sample as input and a 10 dimensional noise vector. This gets fed into a fully connected hidden layer with 256 units and \textit{ReLU} activation. The hidden layer is fully connected to the output layer which uses \textit{sigmoid} as activation function. The size of the output layer is the size of the inputs. After the last layer the output is rounded, thereby transformed into a proper binary feature representation and the constraint of only adding features is enforced.
For the Discriminator we use a simple network with one hidden layer, 256 neurons, \textit{ReLU} activation. The output layer consists of one neuron with \textit{sigmoid} activation. 

The generator and discriminator are trained for 100 epochs. Every epoch the generated \textit{adversarial examples} are tested against a black box detector. At the end the generator with highest misclassification rate on the black box gets used. For the black box detector a model with two hidden layers is used. The first hidden layer has 256 and the second one has 128. Both use \textit{ReLU} as the activation and a dropout of 0.2 during training. The output layer has two neurons and uses \textit{softmax} activation. The black box is trained for 200 epochs using \textit{Adam} with a learning rate of .001, $\beta_1=.9$, $\beta_2=.999$. With this attack we achieve the evasion rates in Table \ref{attacks_malgan}. 

\begin{table}
	\centering
	\caption{Hu et. al \cite{DBLP:journals/corr/HuT17} approach which uses a GAN to create \aexs }
	\begin{tabular}{|c|c|c|c|c|c|c|} 
		\hline
		Dataset 		& acc & acc adv & FNR  & FNR adv  & change & evasion\\
		\hline
		$Drebin_{500}$	& 0.99 & 0.99   & 0.05 & 0.03     & 227.46 & 0.0 \\ 
		\apds 			& 0.93 & 0.70   & 0.12 & 0.43     & 163.90 & 0.77  \\ 
		\hline
	\end{tabular}

	\label{attacks_malgan}
\end{table}

Interestingly when attacking the \Drebinfive{} victim model no evasion at all is achieved. In fact the opposite happens, and all \aexs{} get classified correctly, improving model performance. On \apds{} the attack achieves an evasion rate of 77\%. The attack on the \Drebinfive{} model could possibly be made more effective by using a more complex Discriminator and/or Generator. Also noteworthy is the high modification count. The original approach as presented in \cite{DBLP:journals/corr/HuT17} offers no way of controlling the modification count. It could be possible to add a penalty to the loss of Generator that penalizes large modifications. Another potentially interesting idea to pursue is to use more stable GAN training methods such as Wasserstein GANs \cite{1701.07875} and Improved Wasserstein GANs \cite{DBLP:journals/corr/GulrajaniAADC17}. This could lead to results with a higher evasion rate.

\subsection{Attack Effectiveness}
The effectiveness of an attack can be measured by its ability to create \aexs{} that evade the classifier. The previous experiments have shown that not all attacks work equally well and the victim model that is being attacked also influences the effectiveness. In Table \ref{attack_effectiveness} we summarize how well the different attacks were able to evade the classifiers.      
There are a few interesting things to note. First, the evasion rate for the \Drebinfive{} model is lower for all attacks expect $\text{BGA}^k$ and random inc\_neg. The natural evasion is still lower for the \Drebinfive{} model. $BGA^k$ on \Drebinfive{} achieves 0.0\% increase in evasion. For random inc\_neg the increase in evasion is 7.4\% for \Drebinfive{}. This makes the model trained on the \Drebinfive{} dataset more resistant to adversarial examples than the other models. One possible explanation is that we perform feature selection on the data before training the model, leaving the \aexs{} creation algorithms less to work with. Zhang et al. \cite{7090993} investigate feature selection as a possible defense against evasion attacks.

\begin{table}
	\centering
	\caption{Comparison of the different attacks effectiveness' on all victim models}
\begin{tabular}{m{1.5cm}rrr}
    \toprule
    Attack & evasion ( in \% ) on:  \Drebinfive	& \apds	\\
    \midrule
    natural                     	& 5.6		& 11.7 	\\
    dfgsm\_k                    	& 15.0		& 100.0 \\
    rfgsm\_k                    	& 15.0		& 100.0 \\
    bga\_k                      	& 5.6		& 11.7	\\
    bca\_k                   		& 12.7		& 13.7	\\
    JSMA                        	& 32.8		& 100.0 \\
    random\_inc\_neg            	& 6.0		& 11.7	\\
    dec\_pos                    	& 25.4		& 100.0	\\
    inc\_neg                    	& 32.8		& 100.0	\\
    random\_dec\_pos            	& 24.7		& 100.0 \\
    random\_dec\_pos\_inc\_neg  	& 18.1		& 96.1 	\\
    dec\_pos\_inc\_neg          	& 27.0		& 100.0	\\
    malgan                      	& 0.0		& 77.1	\\
    \bottomrule
\end{tabular}

\label{attack_effectiveness}
\end{table}

\section{Defenses}

\subsection{Distillation}
Distillation was originally introduced as a technique to train smaller models. To achieve this, the smaller model is trained with soft class labels which is the output of the bigger model. Distillation as a defensive technique also trains a second network with soft class labels, but in this case the two networks have the same architecture. The idea is not to make the network smaller but to make it more resilient to adversarial examples by being able to generalize better. Papernot et al. \cite{DBLP:journals/corr/PapernotMWJS15} propose the idea for the image classification domain, \cite{DBLP:journals/corr/GrossePM0M16} and \cite{DBLP:journals/corr/abs-1712-05919} have investigate distillation as a possible defense against \aexs{} in malware detection. In their experiments it did not perform as well on malware as on images. Additionally choosing a temperature T that is too high will actually hurt the performance of the model. We chose T=10 in accordance with the result of \cite{DBLP:journals/corr/abs-1712-05919}. As the table shows, distillation does not actually add any robustness to the network. The impact on accuracy is almost negligible. The undefended network achieves an accuracy of 93\% while the distilled network manages an accuracy of 91.6\%. When training the \Drebinfive{} model with distillation and T=10 the model already has a 1.0 FPR, making evasion attacks superfluous. This may be due to feature selection, the class imbalance, or a combination of the two. 

\subsection{Adversarial Training}

Szegedy et al. \cite{DBLP:journals/corr/SzegedyZSBEGF13} propose a strategy called adversarial training. The idea is that adding adversarial examples to the training data acts as regularization. This improves the generalization capabilities of the model and therefore makes it more resistant against adversarial examples. The idea has been picked up by many different researchers and modified. Kurakin et al. \cite{DBLP:journals/corr/KurakinGB16a} introduce a scalable adversarial training framework based on mini batches. They also find that examples created with single step methods such as FGSM offer greater benefits than iterative approaches. On the other hand, when using adversarial samples for adversarial training  Mossavi et al. \cite{DBLP:journals/corr/Moosavi-Dezfooli15} find that using DeepFool samples increases robustness, whereas using FGSM samples could lead to decreases in robustness. Tramèr et al. \cite{1705.07204} find that models trained with adversarial training are more resistant to white box attacks but, counter intuitively, are still vulnerable to black box attacks.
Multiple authors have looked at adversarial training as a defense in the malware domain. \cite{DBLP:journals/corr/GrossePM0M16} find that the amount of \aexs{} introduced during training needs to be carefully controlled. In their experiments, adding more than 100 \aexs{} during training left the network more vulnerable. Al-Dujaili et al. \cite{DBLP:journals/corr/abs-1801-02950} describe adversarial learning as a saddle point problem. It relies on finding the parameters $\theta$ for a classifier that minimizes some loss function $L$ when assigning class labels $y$ to inputs $x$ taken from a dataset $\mathcal{D}$. Creating an \aex $x_{adv}$, on the other hand, is a maximization problem. Given all binary representations $\mathcal{S}(x) \subseteq \mathcal{X}$ of a malware example $x$ that still preserve the functionality of $x$. We are looking for $\mathcal{S}^*(x) \subseteq \mathcal{S}(x) $ that maximize the adversarial loss.
To create an \aex{} we need to find $\bar{x}$ for which the adversarial loss is maximal. Combining both the learning problem and the adversarial loss gives us the equation for adversarial learning
\begin{equation}
\theta^* \in \arg \min_{\theta \in \mathbb{R}^P } \mathbb{E}_{(x,y)\sim \mathcal{D}} \bigg[ \max_{\bar{x} \in \mathcal{S}(x) } L(\theta,\bar{x},y) \bigg]
\end{equation}
The outer minimization problem can be solved by various different gradient descent algorithms. The inner maximization problem, the maximization of the adversarial loss, can be solved by multiple methods. \cite{DBLP:journals/corr/abs-1801-02950} propose four different approaches, discussed earlier in the attacks section. Theoretically any method to create \aexs{} can be used here. We use $dFGSM_k$ to create \aexs{}. 
The results suggest that adversarial training is very efficient at making the network more robust against \aexs{}. Another possibly beneficial strategy is using \aexs{} from different attacks during training. 

\subsection{Ensembles}

A modification of the ensemble method is proposed by Tramèr et al. \cite{1705.07204} a defense called \textit{Ensemble Adversarial Training}. 
As in normal adversarial training, adversarial examples are included in the dataset. In Ensemble Adversarial Training, the adversarial examples come not only from the model being trained but also from another pretrained model. This training technique enhances the defense against black box attacks. The authors do not consider Adaptive Black-Box attacks which use the output of the target model to enhance their attack. 
\cite{DBLP:journals/corr/abs-1712-05919} looked at ensemble learning to defend malware detectors. 
In our experiments we train three different models and combine them into an ensemble. For ensembles to work it is best to have models that behave differently. For that reason we have chosen three very different architectures. The three architectures are as follows. 1. For our first model we have chosen a very wide but rather flat architecture. It consists of two fully connected hidden layers with 1000 neurons each. 2. The second model was designed to be the opposite of the first model a rather flat but deeper architecture. It consists of eight hidden layers with 64 neurons each. 3. The third model is a rather small model with 2 hidden layers consisting of 256 neurons for the first one and 128 for the second. All models use \textit{ReLU} as activation function for all layers, except the last layer which uses \textit{softmax}. To force the models to be more different from each other we employ heavy dropout. Both hidden layers have a dropout value of 0.5. We train all three models for 200 epochs using the Adam optimizer with our standard settings described earlier in the paper.  We test the robustness of the ensemble of networks with all of our attacks.
Compared to the victim model, the natural evasion of the ensemble is better, but the ensemble does not add anything in terms of adversarial robustness. The ensemble performs worse, when possible, when under attack. This backs up Goodfellow et al. claims that ensemble is not a good defense. A potential idea that could improve the resistance of ensembles in a black box setting is having the different classifiers work on different features. Since binary programs do not represent the feature for a machine learning classifier directly, like images do for example, they need to be extracted first. There are different ways of extracting the features from an application. The two main methods are dynamic and static feature extraction. Dynamic feature extraction collects the features from the application at running time while static extraction does not execute the program and only performs analysis on the compiled code or other available resources. Different models in the ensemble could use different features to train on. This might improve the resilience and make it harder for an attacker to find features to modify. Although Rosenberg et. al \cite{DBLP:journals/corr/RosenbergSRE17} have devised an attack that works with an ensemble of two classifieres where one classifier learns from static and the other learns from the dynamic features.  

\subsection{Random Feature Nullification}

To make the gradient harder for an attacker to compute, Wang et al. \cite{DBLP:journals/corr/WangGZXGL16} introduce a defense called random feature nullification. During training and classification a random set of features gets disabled. This is achieved by choosing a binary vector $I_{p^i}$ and computing the Hadamard-Product of the instance $x_i$ and the vector $I_{p^i}$. The zeros in $I_{p^i}$ are uniformly distributed. The number of zeroes and the distribution are hyperparameters that can be tuned. Increasing the number of features that are nullified will hurt performance, and not disabling enough will have no impact on the attacker's ability to create \aexs{}. It is simple to see that with nullifying all features the input would be a vector of all zeros, making meaningful predictions impossible. Disabling only a very small subset of features, especially on high dimensional data, decreases the chances of hitting features that are important for an attacker. \cite{DBLP:journals/corr/GrossePM0M16} found that they were only modifying 0.0004\% of the features to mislead the classifier. 
The results vary quite a lot, which is to be expected given the random nature of the network. It is possible that averaging could make the model more robust. This raises the question, though, of wheter averaging would hurt the robustness and could be exploited by an attacker. In the end averaging over different runs is analogous to having an ensemble of classifiers with averaged results. As shown earlier an ensemble offers little to no improvement in adversarial robustness.

\begin{table}
	\centering
	\caption{Comparison of the different attacks effectiveness on all victim models. Reported is evasion (in \%). Used defenses: undefended (undef.), Distillation (Dist.), Random Feature Nullification (RFN), Adversarial Training (AT) and Ensembles (Ens.) }
\begin{tabular}{lrrrrr}
    \toprule
    Attack& undef. 	&  Dist. & RFN	& AT 	& Ens.	\\
    \midrule
    natural                     	& 11.7		& 12.0			& 85.6 	& 2.7					& 8.7		\\
    dfgsm\_k                    	& 100.0		& 100.0			& 13.7	& 7.8					& 100.0		\\
    rfgsm\_k                    	& 100.0		& 100.0			& 9.8 	& 7.8					& 100.0		\\
    bga\_k                      	& 11.7		& 100.0			& 72.5	& 3.9					& 100.0		\\
    bca\_k                   		& 13.7		& 31.4			& 68.5	& 3.9					& 17.6		\\
    JSMA                        	& 100.0		& 100.0			& 100.0 & 27.5					& 100.0		\\
    random\_inc\_neg            	& 11.7		& 13.7			& 3.9	& 3.9					& 15.7		\\
    dec\_pos                    	& 100.0		& 100.0			& 96.1	& 100.0					& 100.0		\\
    inc\_neg                    	& 100.0		& 100.0			& 100.0	& 27.5					& 100.0		\\
    random\_dec\_pos            	& 100.0		& 100.0			& 86.3 	& 100.0					& 100.0		\\
    random\_dec\_pos\_inc\_neg  	& 96.1		& 90.2			& 51.0	& 25.5					& 98.0		\\
    dec\_pos\_inc\_neg          	& 100.0		& 100.0			& 100.0	& 25.5					& 100.0		\\
    malgan                      	& 77.1		& 95.8 			& 100.0	& 62.5					& 12.5      \\
    \bottomrule
\end{tabular}

\label{def_effectiveness}
\end{table}

\section{Discussion}

Comparing the different defenses paints a pretty bleak picture. A side by side comparison can be seen in Table \ref{def_effectiveness}. Distillation has almost no positive impact on adversarial robustness and effectively hurts it against certain attacks. Random feature nullification could be potentially beneficial, as it increases the robustness significantly, but in our experiments it also hurts the unattacked performance considerably. While the ensemble proves to be helpful in improving natural evasion, it does not protect against adversarial evasion. It is possible that having classifiers in the ensemble, trained on different feature representations of the instance, would make the system more robust. The only defense that provided significant robustness is adversarial training. We use the saddle point formulation presented in \cite{DBLP:journals/corr/abs-1712-05919} and the results look promising. It improves robustness against all attacks that adhere to our threat model. We also suggest a possible improvement to training process that includes \aexs{} from more than one attack, possibly hardening the classifier further against more attacks. The problem with adversarial training is that it is most efficient against attacks that have been used during the training. Therefore, it does not necessarily provide robustness against new yet unknown attacks. 
Even if a defense does not impact the performance of the classifier significantly, it might  make it more vulnerable to certain \aexs{}  creation methods. This poses the question of whether the defenses we employ today in fact make models more vulnerable to future attacks.

A good defense should ideally be robust against unknown attacks. So far, none of the defenses provide any mathematical hardness guarantees in the way that encryption does in traditional security. There is no guarantee that would make it impossible or at least computationally infeasible for an attacker to create adversarial examples. Not only is the problem of adversarial examples unsolved as shown in \cite{DBLP:journals/corr/CarliniW16a}, defense performance also varies when applied to malware data. This suggests that the defense needs to be adapted to the domain in some way. All of the defenses compared in this paper are domain agnostic. Perhaps an approach tailored more directly to binary data could improve resilience. 

Furthermore, there is currently no special adaptation for binary data in machine learning and neural networks. In image classification, convolutional layers have moved feature extraction into the network itself. In text processing, embedding layers have done something similar. For binary data, however, no such layers exist. Security could be taken into consideration in the design of such layers for malware detection, as well as in feature extraction.
\section{Related Work}

Yuan et al. \cite{DBLP:journals/corr/abs-1712-07107} have compiled a summary of available methods for creating adversarial examples and defense mechanisms. Their work is not limited to malware detection but contains a section on applications to malware detection. They provide an overview of the different approaches and the underlying algorithms but do not provide any numbers which would allow for a comparison based on effectiveness. Carlini et al. \cite{DBLP:journals/corr/CarliniW16a} look closely at ten available defenses and show that none of them are unbeatable. All attacks and defenses evaluated in that paper are in the image domain.
Rosenberg et al. \cite{DBLP:journals/corr/RosenbergSRE17} use recurrent neural networks (RNN) to do malware detection on dynamic features. The authors pair the RNN with an deep neural network (DNN) to learn from the static features. Their goal is to come up with an attack which evades a system that takes advantage of different feature representations. The dynamic features that the RNN learns from are call sequences. The attack on the static classifier works similarly to the attacks described in this paper. The attack on the RNN is based on an attack working on text data \cite{DBLP:journals/corr/PapernotMSH16}. 
The paper also features a comparison of different RNN architectures and the effectiveness of the attack on a given architecture. With a simple RNN being the must vulnerable and BLSTM being the most robust. An interesting discovery the authors make is that the transferability property of \aexs{}, described in \cite{DBLP:journals/corr/SzegedyZSBEGF13}, also holds true for RNNs. 
Hu et al. \cite{DBLP:journals/corr/HuT17a} also work on RNN based malware detection. They propose an attack to leverage against black box classifiers, training their own substitute RNN and using a generative RNN to create \aexs. 
Chen et al. \cite{Chen:2017:SES:3134600.3134636} develop an approach to defend the classifier that relies on feature selection and ensembles. Their feature selection technique is named SecCLS and the ensemble learning technique is called SecENS. By combining both methods, the authors develop a system called SecureDroid. 
In \cite{DBLP:journals/corr/ChenXFHXZ17} Chen et al. propose a secure two phased learning system. During the offline learning phase a classifier is trained and during the online detection phase suspicious false negatives are used in further training the classifier. 
Yang et. al \cite{Yang:2017:MDA:3134600.3134642} propose feature evolution and confusion. These techniques look at how different features evolve over different versions of malware and can be used to create new synthetic malware instances for a classifier to be trained on.

\section{Conclusion}
We studied the space of adversarial examples in malware classification by evaluating twelve attacks and four defenses on different datasets. Our experiments show that attacks work reliably and it is relatively easy and straightforward to create adversarial examples. The defenses, on the other hand, are not as straightforward and might not always be suitable. The right choice of defense depends on many different factors such as dataset, model architecture and more. 
Our results show that most defense mechanisms adapted from image classification, with the exception of adversarial training, are not efficient in the malware domain, and there is a need for approaches tailored to binary data to ensure robustness of the classifiers.

\bibliographystyle{unsrt}
\bibliography{bib}

\end{document}